\documentclass[journal]{IEEEtran}
\usepackage{graphics} 
\usepackage{epsfig}
\usepackage[utf8]{inputenc}
\usepackage[english]{babel}
\usepackage{nomencl}
\makenomenclature

\usepackage{times} 
\usepackage{graphicx}
\usepackage{float}
\usepackage{algorithm}
\usepackage{algorithmic}
\usepackage[algo2e,ruled]{algorithm2e}

\usepackage{tikz}
\usetikzlibrary{shapes,arrows}

\usepackage{verbatim}
\usepackage{amsmath} 
\usepackage{amssymb}  
\usepackage{amsmath}
\usepackage{amssymb}
\usepackage{mathtools}
\usepackage{mathrsfs}

\usepackage{graphicx}              
\usepackage{ulem}
\usepackage{multirow}
\usepackage{amsthm}
\usepackage{booktabs}
\graphicspath{{GAN2019Figs/}}

\title{A Novel GAN-based Fault Diagnosis Approach for Imbalanced Industrial Time Series}
\author{Wenqian Jiang, Cheng Cheng, Beitong Zhou, Guijun Ma and Ye Yuan
\thanks{This work was  supported by the National Natural Science Foundation of China under Grant 91748112 and by the Primary Research \& Development Plan of Jiangsu Province [grant number BE2017002]. (Corresponding author: Prof. Ye Yuan)}
\thanks{Wenqian Jiang is with China-EU Institute for Clean and Renewable Energy, Huazhong University of Science and Technology, Wuhan, China, 430074.}
\thanks{Cheng Cheng, Beitong Zhou, and Ye Yuan are with School of  Artificial Intelligence and Automation, Huazhong University of Science and Technology, Wuhan, China, 430074. Ye Yuan is also with the State Key Lab of Digital Manufacturing Equipment and Technology }
\thanks{Guijun Ma is with School of Mechanical Science and Engineering and the State Key Lab of Digital Manufacturing Equipment and Technology, Huazhong University of Science and Technology, Wuhan, China, 430074.}
}

\begin{document}

\maketitle

\begin{abstract}
This paper proposes a novel fault diagnosis approach based on generative adversarial networks (GAN) for imbalanced industrial time series where normal samples are much larger than failure cases. We combine a well-designed feature extractor with GAN to help train the whole network. Aimed at obtain data distribution and hidden pattern in both original distinguishing features and latent space, the encoder-decoder-encoder three-sub-network is employed in GAN, based on Deep Convolution Generative Adversarial Networks (DCGAN) but without Tanh activation layer and only trained on normal samples. In order to verify the validity and feasibility of our approach, we test it on rolling bearing data from Case Western Reserve University and further verify it on data collected from our laboratory. The results show that our proposed approach can achieve excellent performance in detecting faulty by outputting much larger evaluation scores.
\end{abstract}

\begin{IEEEkeywords}
Fault diagnosis, generative adversarial networks, rolling bearings.
\end{IEEEkeywords}

\section{Introduction}
\label{sec:Intro}
\IEEEPARstart{T}{i}mely and accurate fault diagnosis in industrial systems is of utmost importance. Utilizing acquired measurements and other monitoring information about machine status can help to detect where is damaged or when is about to damage. Thus, fault diagnosis plays a significantly important role in ensuring industrial production is carried out normally and orderly. Taking the characteristics of industrial process data into consideration, there are usually two points of view for fault diagnosis. One is based on the analysis of the failure mechanism which needs one to be familiar with structure of the monitored component, vibration mode, fault performance and so on. The other is based on a ``black box'' pattern where the core algorithm is dedicated to extracting features and pattern recognition. Machine learning, especially the prosperity of deep learning, makes the latter increasingly occupy a pivotal position in industrial fault diagnosis. 

For fault diagnosis in industrial area, the most typical data are physical signals recorded by specific sensors over a duration, such as current and voltage signals, also known as time series data. Using time series for fault diagnosis is always seen as a binary classification problem. At present, researchers prefer to extract useful features perfectly representing time series, and then adopt classification algorithm for fault detection based on these distinguishing features. On the one hand,  feature-based models aimed at different datasets can effectively promote smooth complement of fault detection algorithm. On the other hand, because of the advancements in deep learning to extract rich hierarchical features and achieve good performance for classifying, deep learning models for time series analysis has been well studied. However, using deep learning models for time series anomalies faces two difficulties  :  1) A large number of labeled datasets are essential for deep learning model during training stage. But in many practical industrial systems, samples from abnormal operating condition are often of insufficient data sizes. The imbalance of the positive and negative samples will cause the prediction results biased towards positive in testing stage. In addition, when the equipment runs normally followed by abnormally for a period of time, it is difficult to clearly find out the starting point of abnormality from collected datasets. Thus, it is impossible to clearly label the data, which will also have a very adverse impact on the training of the model;   2) Time series datasets tend to be very large. In general, diversified sensors are responsible for collecting abundant information, and each sensor typically records data continuously at relatively high frequencies in time. If one directly feeds these data  into deep neural network for training, not only is calculation huge but also training effect is minimal. Before using the neural network, how to effectively preprocess these huge datasets and extract useful features is problem that are expected to be solved.

To solve the imbalanced industrial data for fault diagnosis and preprocess the huge amount of  time series before training the model, motivated by \cite{akcay2018ganomaly, akccay2019skip}, we propose a novel GAN\cite{goodfellow2014generative}-based approach combing the advantages of feature extractor and GAN. The main contributions of this paper are as follows:
\begin{enumerate}
\item For the univariate time series in the industrial field, a fault detection algorithm based on GAN  is proposed for the first time. We add a feature extractor specific for industrial time series which is able to present the unique feature of a period and at the same time reduce dimension and computing time before the data is feed into our fault detector. We test our idea on the benchmark dataset ( rolling bearing data from Case Western Reserve University) at first and then validate on datasets collected from our own laboratory, the results show that our algorithm achieves good performance for fault diagnosis on above datasets.
\item We use time series as the input to our algorithm drawn only from normal samples during the training process, which is very helpful for solving the problem of less data of fault samples as a common scenario in the industrial field. 
\end{enumerate}

This paper is organized as follows. Section \ref{sec:Related_Works} reviews related works including feature-based models and deep learning models for fault diagnosis with time series. Section \ref{sec:Approach} proposes our fault diagnosis framework based on GAN. Experiment setup and results are given in Section \ref{sec:Setup} and Section \ref{sec:Result}. Finally, conclusion and future work are drawn in Section \ref{sec:conclusion}.

\section{Related Works}
\label{sec:Related_Works}
Fault diagnosis has long been a question of great interest in industrial process systems. A considerable amount of work has been published to propose efficient theory and algorithm for detecting fault in industrial time series data. Our review is primarily focused on classic feature-based models and several efficient deep learning models.

Feature-based models aim to extract time domain features \cite{nanopoulos2001feature, wang2006characteristic}, frequency domain features(FFT) or time-frequency domain features (wavelet analysis) followed by traditional classification method (principal components analysis, SVM , random forest and so on). In contrast, deep learning models show more and more outstanding performance than featured-based. 

Many anomaly detection techniques used in time series data have been well developed, such as Long Short Term Memory networks in \cite{taylor2016anomaly} Recurrent neural networks in \cite{guo2016robust}, Convolution neural networks in \cite{kanarachos2017detecting}, Autoencoders in \cite{veeramachaneni2016ai}. Recently, Li et al. proposed a novel GAN-based Anomaly Detection (GANAD) method combining GAN with LSTM-RNN to detect anomalies on multivariate time series in \cite{li2018anomaly}. Lim et al. first put forward a data augmentation technique focused on improving performance in unsupervised anomaly detection based on GAN \cite{lim2018doping}.

As can be seen from above, more recent attention in the literature has been focused on the provision of adversarial training, especially on GAN. GAN, viewed as an unsupervised machine learning algorithm, since initially introduced by Goodfellow et al. in 2014, has achieved outstanding application effects in the field of image recognition. Based on GAN, there has been emerged various kinds of adversarial algorithm. For further details, we refer the interested reader to a website which gives a very comprehensive summary of GAN and its variants [website: https://github.com/hindupuravinash/the-gan-zoo]. Last year, based on GAN, a generic anomaly detection architecture called GANomaly put forward by Samet et al. in \cite{akcay2018ganomaly} shows superiority and efficacy compared with previous state-of-the-art approaches over several benchmark image datasets, which gives us an inspiration for fault diagnosis in industrial area. To explain our approach thoroughly in next part, we will briefly introduce GANomaly.

As we all know, GAN consists of two networks (a generator and a discriminator) competing with each other during training such that the former tries to generate an image similar to the real, while the latter determines whether the image is real or generated from the generator. Based on GAN, Samet et al. employ encoder-decoder-encoder sub-networks in the generator network to train a semi-supervised network. They build the network architecture by using DCGAN and employ three loss functions in generator to capture distinguishing features in both input images and latent space. They first proposed a training algorithm for no-negative samples and achieved state-of-the-art performance for anomaly detecting in some image benchmark datasets.

\section{Our Approach}
\label{sec:Approach}

              \begin{figure}[htb!]
              \centering
              \includegraphics[width=1\columnwidth]{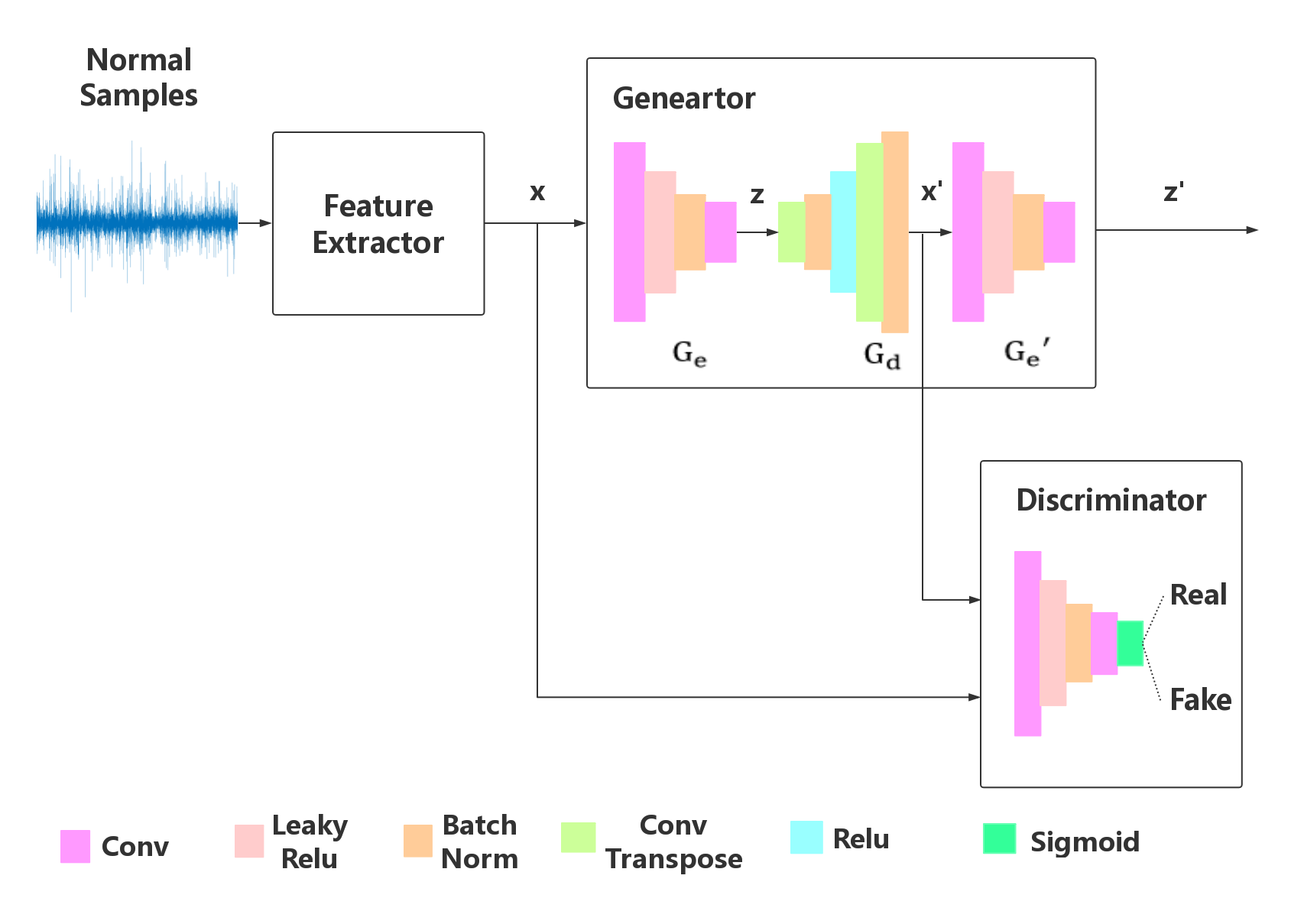}\\    
              \caption{Overview of our proposed training procedure}
              \label{fig:Approach description}
              \end{figure}

Motivated by \cite{akcay2018ganomaly} and \cite{akccay2019skip}, aimed at univariate time series data in industrial area, we propose a new network trained only on normal samples aimed at detecting fault in time series dataset from industrial area.  We adopt the similar encoder-decoder-encoder three-sub-networks in generator, but with a different network architecture. In addition, we add a feature extractor before generator to help train the whole network. During training, we first extract features of the univariate time series data, and then obtain data distribution and potential representative mode of normal samples by our designed faulty detector from extracted features. Finally, diagnose faults or anomalies by outputting higher scores in test samples. We will explain our algorithm in detail.

 Problem definition: given an univariate time series dataset $D = \left\{ X_1,X_2,\ldots, X_n \right\}$,where$ X_i = \left\{ x_1,x_2,\ldots, x_t \right\}$ representing data recorded by one sensor in a period of time, we need to analyze whether each sample in D is normal, which means the algorithm should output 0(normal) or 1(faulty) corresponding to each sample.
 
 In our algorithm, input dataset D are just normal samples during training stage. If we just take normal samples into consideration, the generator will explore and generate possible representation mode of normal data distribution. Once faulty samples are feed into our anomaly detector, the generator will encode and decode samples as the normal, leading to large deviation from the original elements and these clear differences shown on distance between the original and generated will help us find faulty. After training, the test dataset will include both normal and abnormal classes.
 
 As is shown in Figure \ref{fig:Approach description}, the network structure of our algorithm consists of three parts: feature extractor, generator and discriminator. 
              
In order to analyze whether the equipment is faulty, it is usually necessary to record data during continuous hours as a sample point. A sample point usually includes tens of thousands or even hundreds of thousands of data. Apparently, it’s impossible to directly feed such huge amount of data into training networks. Therefore, we need a feature extractor to reduce dimension of samples. In order to maximize the retention of feature information, each sample is subsampled which is equal in length firstly and then subsamples are used to extract features. There are usually two ways utilized to extract characterization information in feature extractor: artificial extraction and neural network. Manual extraction is a purposeful method, which means that researchers have known what feature information will be acquired before inputting samples, such as maximum value, minimum value, variance, steepness, frequency, skewness and so on mentioned in some conventional literatures on time series analysis. In contrast, feature extraction by neural network is purpose-free, where it is not known what the final output features will be. It is a black box mode to extract specific pattern for a specific dataset. Clearly, both methods have their own advantages and disadvantages. Different feature extraction methods can be considered for different datasets. Although neural network, especially deep neural network can automatically capture useful information about the task so that heavy crafting on data processing and feature engineering will be avoided, such non-parametric learning algorithms require a lot more data to train and suffer seriously from overfitting. We believe that elaborately designed neural networks feed with carefully selected features show good performance in plenty of time series tasks. In addition, neutral networks have shown in our generator and discriminator, so for univariate time series data studied in our paper, we just employ artificial extractor. That is not to say feature extractor can be chosen randomly, because we find if we choose more relevant information about faulty diagnosis (e.g. some important information from physical model and analysis), the performance will become more amazing.

For generator, the two encoder learn to acquire input samples representation and generated samples representation respectively and the decoder tries to reconstruct input data at the same time. The whole process is as follows: data X from feature extractor is feed into $G_e$, whose architecture consists of convolutional layers followed by batch-norm and leaky ReLU activation. $G_e$ downscales X into latent representation z, which is the material used by $G_d$ to recreate the input samples. $G_d$ adopts convolutional transpose layers, ReLU activation and batch-norm. Unlike conventional DCGAN, the last layer of Gd does not employ Tanh activation function to scale data in [ -1,1]. The architecture of last encoder $G_e'$ is the same as $G_e$ with different parametrization, and the output $Z'$ is the same as Z in terms of data dimension. The generator guarantees that not only the characteristics of the input samples, but also the pattern of the latent space can be learned at the same time.

The discriminator adopts the standard discriminator network introduced in DCGAN \cite{radford2015unsupervised}, which is used to distinguish whether input data is real or generated.
Having defined our overall network architecture, we now continue to discuss how we define loss function for learning.

In the training phase, because only the normal samples are considered, the three sub-networks of the generator only obtain normal pattern. But in the testing phase, the generator still processes fault samples according to the model acquired during training stage, which means that outputs of the decoder and the second encoder will be similar to the outputs of normal samples, inevitably deviating from the input fault samples and latent vector from the first encoder respectively, so that help us identify fault parts. 

\noindent $\bf{Fraud \ Loss.}$  We base the computation of the fraud loss $L_f$ on the discriminator output by feeding the generated sample into the discriminator, and the formula is as follows: 
\begin{equation}
L_f=\sigma(D(G(z)),\alpha)
\end{equation}
where $\sigma$ is the binary cross entropy loss function. To fool discriminator, we define the fraud loss of generated samples during adversarial training, with D(G(z)) and targets $\alpha =1$.

Fraud loss is aimed to induce the discriminator to judge generated samples from generator as real samples. It is not enough for the generator to learn potential patterns under normal samples and to reconstruct generated samples as realistically as possible, so we define apparent loss measuring $L_1$ distance between the original and the fake samples:

\noindent $\bf{Apparent \  Loss.}$  We base the computation of the fraud loss $L_a$ on the discriminator output by feeding the generated sample into the discriminator, and the formula is as follows: 
\begin{equation}
L_a=\left\|x-x'\right\|
\end{equation}

\noindent $\bf{Latent \  Loss.}$  In addition to fraud loss and apparent loss, we also define latent loss to minimize the distance between the latent representation of real samples and the encoded bottleneck features of generated samples. This loss can help to learn latent representation both in real and fake examples.
\begin{equation}
L_l=\left\|z-z'\right\|
\end{equation}

In summary, the loss function of the generator consists of three parts:
\begin{equation}
L=\omega_f*L_f+\omega_a*L_a+\omega_l*L_l
\end{equation}
For discriminator, feature matching loss is adopted for adversarial learning, which is proposed by Salimans et al. \cite{salimans2016improved} to reduce the instability of GAN training.
\begin{equation}
L_d=f(x)-f(G(x))
\end{equation}

In the testing phase, our model use latent loss and apparent  for scoring the abnormality of a given subsample. Anomaly score is defined as
\begin{equation}
A(x) = L_a+L_l
\end{equation}

\section{Experimental Setup}
\label{sec:Setup}

In order to evaluate feasibility and effectiveness, we first test our algorithm on rolling bearing data from Case Western Reserve University(CWRU), and then further validate it by using rolling bear dataset collected from our laboratory. The two datasets are as follows.\vspace{3ex}

\noindent $\bf{Rolling\ bearing\ data\ from\ CWRU}$  It is a bearing fault diagnosis dataset measuring vibration signal at locations near to and remote from the motor bearings by using accelerometer. Motor bearings were seeded with faults using electro-discharge machining (EDM). Faults ranging from 0.007 inches to 0.040 inches in diameter were introduced separately at the inner raceway, rolling element (i.e. ball) and outer raceway. 
[website: http://csegroups.case.edu/bearingdatacenter/home]\vspace{3ex}

\noindent $\bf{Rolling\ bearing\ data \ from\ our\ laboratory \ and \ Jia-} $\\
\noindent $\bf{ngnan\ University}$  This dataset is similar to that from CWRU, but only using the bearing of 14-mil fault diameter. We record voltage signals from the motor by a Hall sensor (sampling frequency is 50Hz) considering four conditions including normal condition, faulty condition with fault at rolling elements, faulty condition with fault at outer race, and faulty condition with fault at inner race and for each operational condition we experiment on three bearings. 

The procedure for train and test for the above datasets is as follows: We divide normal samples into 80\% and 20\% as training set and test set respectively. In the training stage, only normal samples are considered while the fault sample is included in testing phase. For rolling bearing data from CWRU, considering we just want to test if our algorithm is feasible, we don't bother to design feature extractor. We just subsample(size is 3136) drive end accelerometer signal in normal dataset, and then input subsamples to train our anomaly detector. For rolling bearing data from our lab, we carefully adopt sixteen distinguishing features consist of maximum value, minimum value, average value, standard deviation, peak to peak value, average amplitude, root mean square value, skewness value, waveform indicator, pulse indicator, twist index, peak indicator, margin indicator, kurtosis index, square root amplitude and so on for feature extractor. After validation on our dataset, to show the superiority of our network architecture, We compare our method against another network called bidirectional generative adversarial networks (BiGANs) proposed by \cite{donahue2016adversarial}, because BiGAN based on GAN shows excellent performance on anomaly detection in image field.

We implement our approach in PyTorch \cite{paszke2017automatic} by optimizing the networks using Adam \cite{kingma2014adam} with an initial learning rate 0.001, and momentums 0.5, 0.999. We train the model for 20 epochs for both datasets.

\section{Results}
\label{sec:Result}

\begin{figure}[htb!]
              \centering
              \includegraphics[width=1\columnwidth]{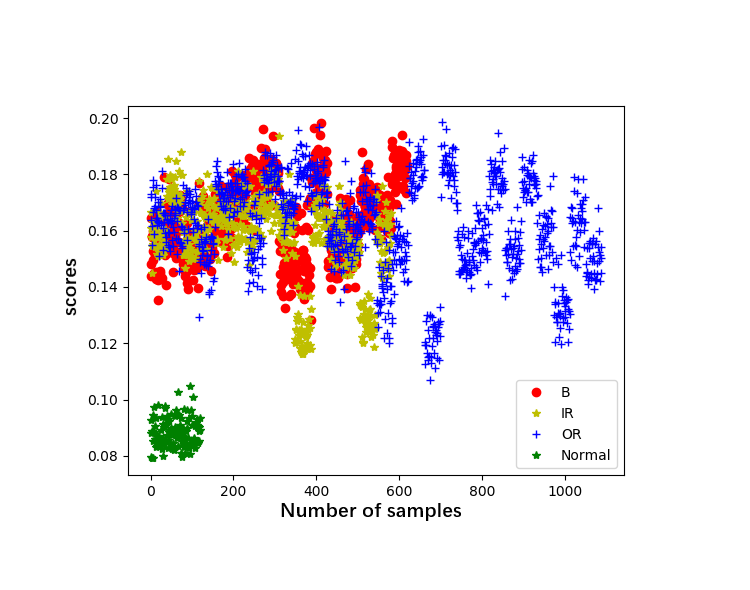}\\    
              \caption{Binary classification on dataset from CWRU}
              \label{fig:cwru1}
              \end{figure}

\begin{figure}[htb!]
              \centering
              \includegraphics[width=1\columnwidth]{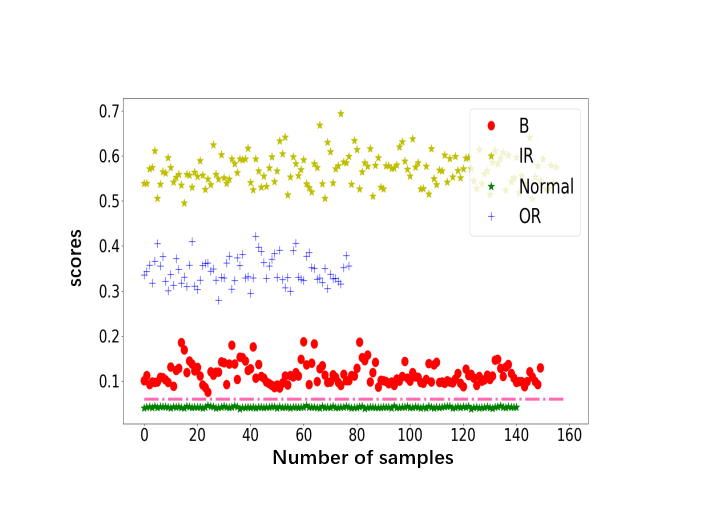}\\    
              \caption{Different types of faulty on  dataset from CWRU }
              \label{fig:cwru2}
              \end{figure}

For both datasets, after training according to the above parameters, the evaluation scores of normal samples and abnormal samples output respectively on the test set. From  Figure\ref{fig:cwru1} , we can make a clear judgment on the failure of the sample data by the level of the score: high score means high possibility of abnormality and vice versa. In addition, we just adjust the weighted factor in general loss, we find that the scores acquired can help us classify different types of faults Figure\ref{fig:cwru2}.

\begin{figure}[htb!]
              \centering 
              \includegraphics[width=1\columnwidth]{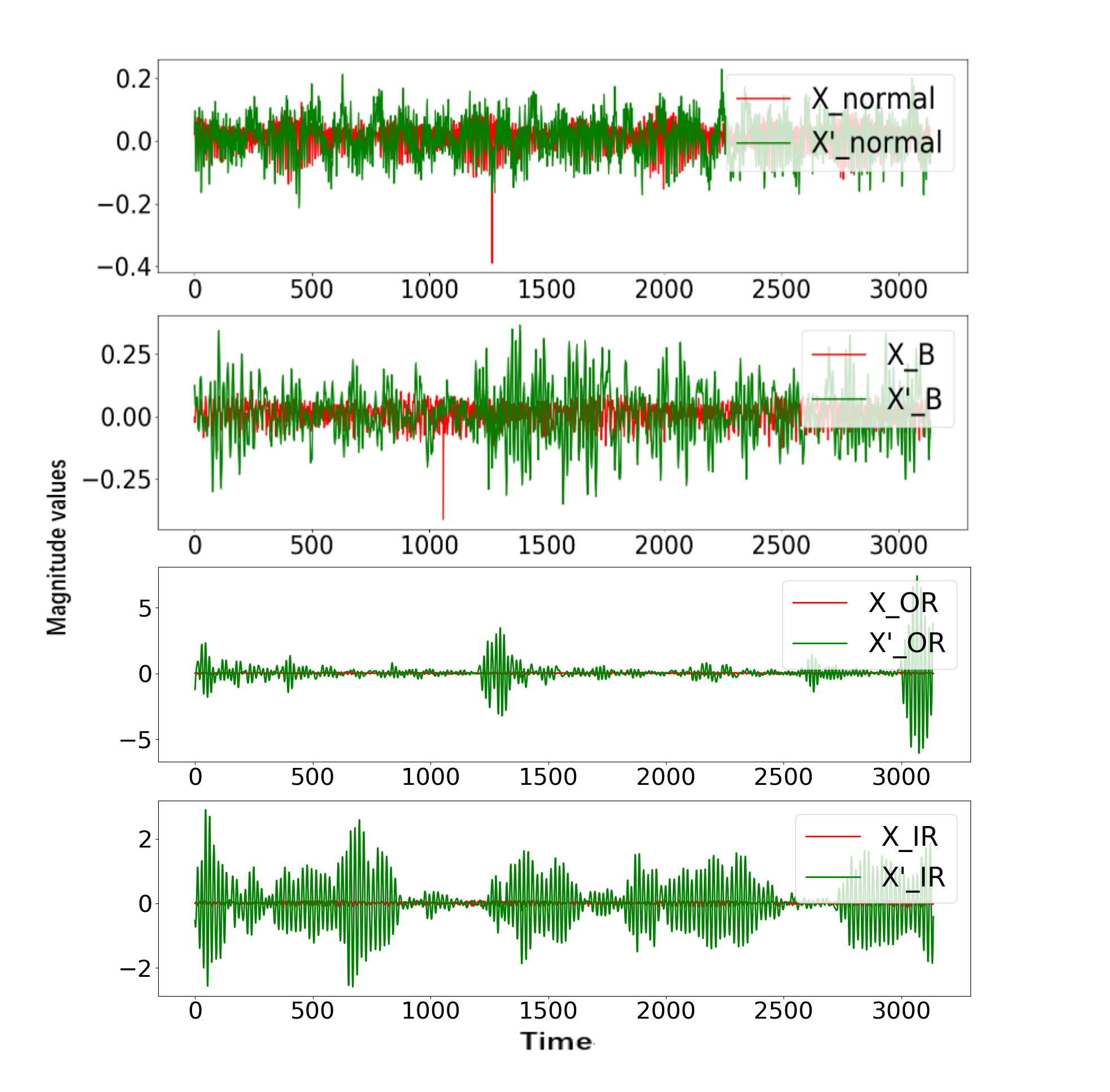}     
              \caption{Comparison between raw  and re-engineered samples on  dataset from CWRU}
              \label{fig:X}
              \end{figure}

\begin{figure}[htb!]
              \centering 
              \includegraphics[width=1\columnwidth]{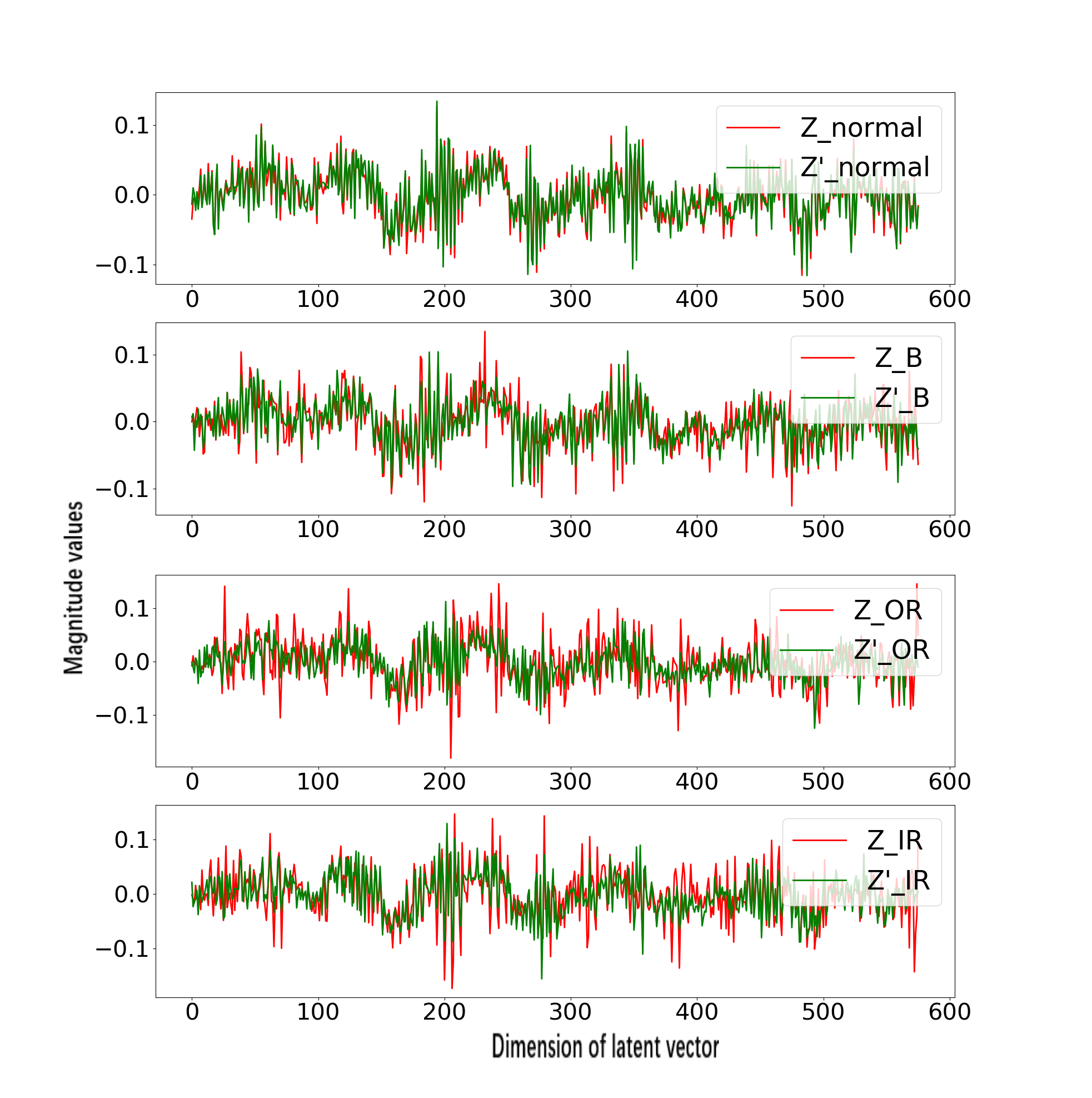}\\   
              \caption{ Latent vectors of raw  and re-engineered samples on  dataset from CWRU}
              \label{fig:Z}
              \end{figure}

We select a normal sample and a fault sample in CWRU dataset randomly, and visualize the two samples on the original sample and the reconstructed sample and the latent representation between them. As is shown in Figure\ref{fig:X} and Foigure\ref{fig:Z}, whether the comparison of raw data with re-engineered data, or the potential space comparison between them, the fault samples are significantly larger than the normal sample. This explains intuitively why the algorithm we proposed is very effective for fault detecting in industrial univariate time series.
                           
\begin{figure}[htb!]
              \centering 
              \includegraphics[width=1\columnwidth]{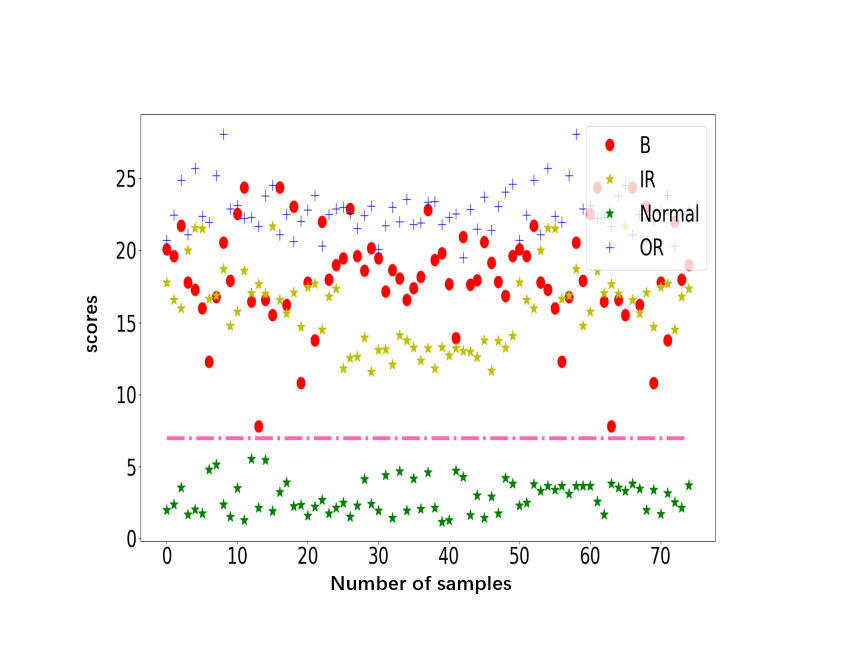}\\    
              \caption{Fault diagnosis performance on dataset from our lab}
              \label{fig:jiangnan_clas}
              \end{figure}

\begin{figure}[htb!]
              \centering 
              \includegraphics[width=1\columnwidth]{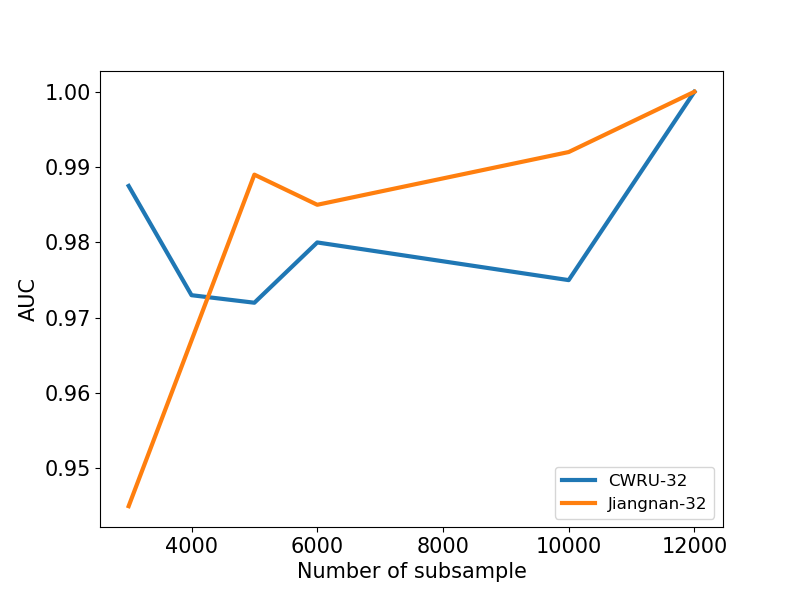}\\   
              \caption{Overall performance of our model based on varying size of the subsample }
              \label{fig:subsample}
              \end{figure}

\begin{figure}[htb!]
              \centering
              \includegraphics[width=1\columnwidth]{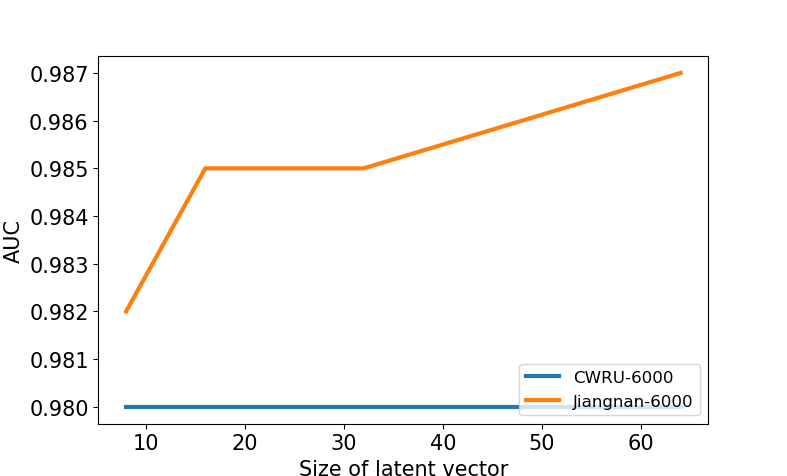}\\    
              \caption{Impact of the size of latent vector on the overall performance}
              \label{fig:latent}
              \end{figure}

 After testing on CWRU, based on rolling bearing data from our lab, we also get excellent performance just as shown in Figur \ref{fig:jiangnan_clas}. In addition, we explore how the choice of hyper-parameters ultimately affect the overall performance of the model. In Figur \ref {fig:subsample}, We see that the optimal performance is achieved when the length of the subsample is 12000 both for two datasets. Considering the sampling frequency is 50Hz for data collected in our lab, we are able to infer that the potential pattern of this dataset is hidden in the data collected every 4 minutes. According to Figure\ref{fig:latent}, we can conclude that when the size of the latent representation is 64, the model will achieve the highest accuracy for our dataset, but the size of latent representation does not  make an effect on final accuracy of data from CWRU.              
              
To further validate effectiveness of our anomaly detector, we make a comparison between our architecture and BiGAN network on different sizes of subsamples. The results show that our algorithm is almost stable on different sizes of subsamples and achieve higher accuracy than BiGAN.

             \begin{figure}[htb!]
              \centering
              \includegraphics[width=1\columnwidth]{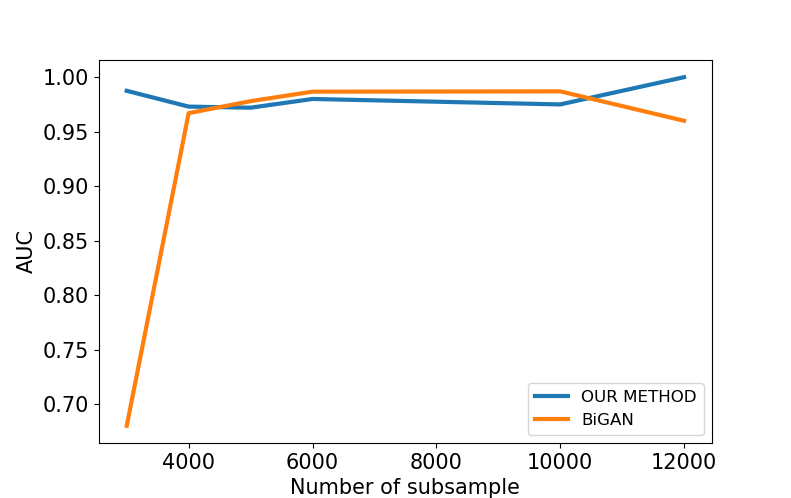}\\    
              \caption{Comparison between BiGANs and our method based on different sizes of subsample }
              \label{fig:bigan-ganomal}
              \end{figure}

\section{Conclusion}
\label{sec:conclusion}
Aimed at imbalanced industrial time series datasets, we put forward an innovative architecture where just normal samples are considered for training to achieve superior fault diagnosis performance. We elaborately design a feature extractor before fault detector based on data characteristics for specific datasets, and an encoder-decoder-encoder generator guarantee that code and reconstruct latent pattern of normal samples and detect the existence of abnormal samples by outputting a large deviation score. Future work should consider more about feature extractor, because different recorded signals often possess different feature modes. In addition, how to combine data information between different dimensions of multivariate time series in the algorithm to achieve better diagnostic effects is also worthy of well studied. 




\begin{thebibliography}{10}
\providecommand{\url}[1]{#1}
\csname url@samestyle\endcsname
\providecommand{\newblock}{\relax}
\providecommand{\bibinfo}[2]{#2}
\providecommand{\BIBentrySTDinterwordspacing}{\spaceskip=0pt\relax}
\providecommand{\BIBentryALTinterwordstretchfactor}{4}
\providecommand{\BIBentryALTinterwordspacing}{\spaceskip=\fontdimen2\font plus
\BIBentryALTinterwordstretchfactor\fontdimen3\font minus
  \fontdimen4\font\relax}
\providecommand{\BIBforeignlanguage}[2]{{%
\expandafter\ifx\csname l@#1\endcsname\relax
\typeout{** WARNING: IEEEtran.bst: No hyphenation pattern has been}%
\typeout{** loaded for the language `#1'. Using the pattern for}%
\typeout{** the default language instead.}%
\else
\language=\csname l@#1\endcsname
\fi
#2}}
\providecommand{\BIBdecl}{\relax}
\BIBdecl

\bibitem{akcay2018ganomaly}
S.~Akcay, A.~Atapour-Abarghouei, and T.~P. Breckon, ``Ganomaly: Semi-supervised
  anomaly detection via adversarial training,'' \emph{arXiv preprint
  arXiv:1805.06725}, 2018.

\bibitem{akccay2019skip}
S.~Ak{\c{c}}ay, A.~Atapour-Abarghouei, and T.~P. Breckon, ``Skip-ganomaly: Skip
  connected and adversarially trained encoder-decoder anomaly detection,''
  \emph{arXiv preprint arXiv:1901.08954}, 2019.

\bibitem{goodfellow2014generative}
I.~Goodfellow, J.~Pouget-Abadie, M.~Mirza, B.~Xu, D.~Warde-Farley, S.~Ozair,
  A.~Courville, and Y.~Bengio, ``Generative adversarial nets,'' in
  \emph{Advances in neural information processing systems}, 2014, pp.
  2672--2680.

\bibitem{nanopoulos2001feature}
A.~Nanopoulos, R.~Alcock, and Y.~Manolopoulos, ``Feature-based classification
  of time-series data,'' \emph{International Journal of Computer Research},
  vol.~10, no.~3, pp. 49--61, 2001.

\bibitem{wang2006characteristic}
X.~Wang, K.~Smith, and R.~Hyndman, ``Characteristic-based clustering for time
  series data,'' \emph{Data mining and knowledge Discovery}, vol.~13, no.~3,
  pp. 335--364, 2006.

\bibitem{taylor2016anomaly}
A.~Taylor, S.~Leblanc, and N.~Japkowicz, ``Anomaly detection in automobile
  control network data with long short-term memory networks,'' in \emph{2016
  IEEE International Conference on Data Science and Advanced Analytics
  (DSAA)}.\hskip 1em plus 0.5em minus 0.4em\relax IEEE, 2016, pp. 130--139.

\bibitem{guo2016robust}
T.~Guo, Z.~Xu, X.~Yao, H.~Chen, K.~Aberer, and K.~Funaya, ``Robust online time
  series prediction with recurrent neural networks,'' in \emph{2016 IEEE
  International Conference on Data Science and Advanced Analytics
  (DSAA)}.\hskip 1em plus 0.5em minus 0.4em\relax Ieee, 2016, pp. 816--825.

\bibitem{kanarachos2017detecting}
S.~Kanarachos, S.-R.~G. Christopoulos, A.~Chroneos, and M.~E. Fitzpatrick,
  ``Detecting anomalies in time series data via a deep learning algorithm
  combining wavelets, neural networks and hilbert transform,'' \emph{Expert
  Systems with Applications}, vol.~85, pp. 292--304, 2017.

\bibitem{veeramachaneni2016ai}
K.~Veeramachaneni, I.~Arnaldo, V.~Korrapati, C.~Bassias, and K.~Li, ``Ai\^{} 2:
  training a big data machine to defend,'' in \emph{2016 IEEE 2nd International
  Conference on Big Data Security on Cloud (BigDataSecurity), IEEE
  International Conference on High Performance and Smart Computing (HPSC), and
  IEEE International Conference on Intelligent Data and Security (IDS)}.\hskip
  1em plus 0.5em minus 0.4em\relax IEEE, 2016, pp. 49--54.

\bibitem{li2018anomaly}
D.~Li, D.~Chen, J.~Goh, and S.-k. Ng, ``Anomaly detection with generative
  adversarial networks for multivariate time series,'' \emph{arXiv preprint
  arXiv:1809.04758}, 2018.

\bibitem{lim2018doping}
S.~K. Lim, Y.~Loo, N.-T. Tran, N.-M. Cheung, G.~Roig, and Y.~Elovici, ``Doping:
  Generative data augmentation for unsupervised anomaly detection with gan,''
  in \emph{2018 IEEE International Conference on Data Mining (ICDM)}.\hskip 1em
  plus 0.5em minus 0.4em\relax IEEE, 2018, pp. 1122--1127.

\bibitem{radford2015unsupervised}
A.~Radford, L.~Metz, and S.~Chintala, ``Unsupervised representation learning
  with deep convolutional generative adversarial networks,'' \emph{arXiv
  preprint arXiv:1511.06434}, 2015.

\bibitem{salimans2016improved}
T.~Salimans, I.~Goodfellow, W.~Zaremba, V.~Cheung, A.~Radford, and X.~Chen,
  ``Improved techniques for training gans,'' in \emph{Advances in neural
  information processing systems}, 2016, pp. 2234--2242.

\bibitem{donahue2016adversarial}
J.~Donahue, P.~Kr{\"a}henb{\"u}hl, and T.~Darrell, ``Adversarial feature
  learning,'' \emph{arXiv preprint arXiv:1605.09782}, 2016.

\bibitem{paszke2017automatic}
A.~Paszke, S.~Gross, S.~Chintala, G.~Chanan, E.~Yang, Z.~DeVito, Z.~Lin,
  A.~Desmaison, L.~Antiga, and A.~Lerer, ``Automatic differentiation in
  pytorch,'' 2017.

\bibitem{kingma2014adam}
D.~P. Kingma and J.~Ba, ``Adam: A method for stochastic optimization,''
  \emph{arXiv preprint arXiv:1412.6980}, 2014.

\end{thebibliography}
\end{document}